\documentclass{article}
\usepackage{spconf,amsmath,graphicx}

\usepackage{xcolor}
\usepackage{url}

\usepackage[utf8]{inputenc} 

\title{Refacing: reconstructing anonymized facial features using GANs}

\name{David Abramian$^{a,b}$ \qquad Anders Eklund$^{a,b,c}$ \thanks{This study was supported by Swedish research council grant 2017-04889. Funding was also provided by the Center for Industrial Information Technology (CENIIT) at Linköping University, and the Knut and Alice Wallenberg foundation project ”Seeing organ function”. We thank the Biomedical Image Analysis Group at Imperial College, London, for for sharing the IXI dataset. The Nvidia Corporation, which donated the Nvidia Titan X Pascal graphics card used to train the GANs, is also acknowledged. 
 }} 

\address{$^{a}$ Division of Medical Informatics, Department of Biomedical Engineering, \\ 
    $^{b}$ Center for Medical Image Science and Visualization (CMIV),  \\
    $^{c}$ Division of Statistics and Machine Learning, Department of Computer and Information Science, \\ Linköping University, Linköping, Sweden }

\begin{document}
%

\maketitle
\begin{abstract}
Anonymization of medical images is necessary for protecting the identity of the test subjects, and is therefore an essential step in data sharing. However, recent developments in deep learning may raise the bar on the amount of distortion that needs to be applied to guarantee anonymity. To test such possibilities, we have applied the novel CycleGAN unsupervised image-to-image translation framework on sagittal slices of T1 MR images, in order to reconstruct facial features from anonymized data. We applied the CycleGAN framework on both face-blurred and face-removed images. Our results show that face blurring may not provide adequate protection against malicious attempts at identifying the subjects, while face removal provides more robust anonymization, but is still partially reversible.
\end{abstract}

\section{Introduction}
Anonymization is an important topic in medical imaging and data sharing, to guarantee privacy for the test subjects. This is especially important for neuroimaging~\cite{Poldrack2014}, where head volumes are collected, and for subjects with a specific disease. Furthermore, the General Data Protection Regulation (GDPR) often requires anonymization. Virtually all data sharing initiatives in the neuroimaging field therefore remove facial features from MRI volumes before they are shared with the community. At least two techniques are currently being used: removing all facial features (e.g. using FreeSurfer \cite{Bischoff2007}) or blurring the face \cite{Milchenko2013}. Face removal was used in the 1000 Functional Connectomes Project~\cite{Biswal2010}, while face blurring was used in the Human Connectome Project~\cite{VanEssen2013}.

Deep learning has been extensively used for medical imaging \cite{Litjens2017,Greenspan2016}, as these new methods in many cases provide superior performance compared to traditional image processing algorithms. In particular, generative adversarial networks (GANs) \cite{Goodfellow2014, Isola2017image, Zhu2017} have recently become a very popular tool for a multitude of tasks, such as realistic image synthesis, denoising, domain translation, and superresolution \cite{Yi2018,Kazeminia2018}. A conditional GAN can for example be used to generate CT images from MRI \cite{nie2017medical, wolterink2017deep}, PET images from MRI \cite{wei2018learning} or T1-weighted MR images from T2-weighted images \cite{dar2018image, Welander2018}. 

New machine learning techniques, in combination with fast computing, have provided great benefits for the medical imaging field. However, these techniques have also opened the door to certain malicious applications. In this work, we attempt to highlight this problem by demonstrating that a
GAN can be used to restore facial features of  anonymized T1-weighted images.  Our code is available at \url{https://github.com/DavidAbramian/refacing}.

\section{Data}
The data used was obtained from the IXI dataset \cite{IXI}, a multi-site MRI dataset including T1, T2, PD, MRA and diffusion data from 581 subjects. In this work we employ only the T1 images, which are provided without any facial anonymization. The images have also not been coregistered or normalized to a common space. Table~\ref{tab:IXI} provides more details about the composition of the IXI dataset and in particular the T1 images.

\begin{table*}[htb]
    \centering
    \begin{tabular}{|c|c|c|c|c|c|}
        \hline
        Location & Scanner & Num. subjects & Image dim. (vox.) & Vox. size (mm) \\
        \hline 
        Guy's Hospital & Philips Gyroscan Intera 1.5T & $322$ & $150\times256\times256$ & $1.2\times0.938\times0.938$ \\
        Hammersmith Hospital & Philips Intera 3T & $185$ & $150\times256\times256$ & $1.2\times0.938\times0.938$ \\
        Institute of Psychiatry & GE 1.5T & $74$ & $146\times256\times256$ & $1.2\times0.938\times0.938$ \\
        \hline
    \end{tabular}
    \caption{Composition of the IXI dataset.}
    \label{tab:IXI}
\end{table*}

\section{Methods}
\subsection{Anonymization}
Two different anonymization procedures were applied to the data. The first was the mask\_face software \cite{Milchenko2013}, which applies blurring to the facial surface while conserving the structure beneath the face. The second is the mri\_deface function from the FreeSurfer package \cite{Bischoff2007}, which zeroes out all the voxels from the subject's face, including deeper facial structures. 

\subsection{GAN model}
We employed the CycleGAN unsupervised image-to-image translation framework \cite{Zhu2017} to reconstruct facial features from anonymized data. CycleGAN is a generative adversarial network which employs two generators and two discriminators, all of which are convolutional neural networks, to simultaneously learn the mappings between the two domains $A$ and $B$. Because the data is unpaired, the problem of finding a mapping between two domains is underdetermined. To counteract this, CycleGAN employs a cycle consistency constraint that requires that data converted to another domain and back be as close to the original as possible.

We used an implementation of 2D CycleGAN previously developed in our group \cite{Welander2018}, based on the Keras API \cite{Chollet2015}. The model is trained to transform images between the anonymized and the original domains. The generators in the GAN have $24$ convolutional layers, while the discriminators have $5$.

\subsection{Training}
The model was trained using individual head slices. To generate the training data, $21$ sagittal slices were extracted from each subject. This was done for the original dataset as well as for both anonymized versions. Each slice was normalized with the $99.5$ percentile value of its corresponding original volume. All the images were of size $256\times256$ pixels. 

Two different sets of images were used for training the CycleGAN: the first included subjects scanned at Guy's Hospital, and the second included subjects scanned at all three locations. In the first case, $6300$ images ($300$ subjects) were used for training and $462$ ($22$ subjects) for testing, while in the second case $10500$ images ($500$ subjects in total; $284$, $151$, and $65$ from each site respectively) were used for training and $1701$ ($81$ subjects in total; $38$, $34$, and $9$ from each site respectively) were used for testing.

In both cases the training was performed for $200$ epochs, with linear decay of the learning rate applied during the second half of the training. The training time for the Guy's data was about $2$ days, while for the whole dataset it was close to $4$ days on an Nvidia Titan X Pascal graphics card.

\subsection{Evaluation}
Qualitative evaluation of the results was performed by visual comparison of the original and reconstructed images. Quantitative results are provided in the form of correlation coefficients and structural similarity indices (SSIMs) between the original and the reconstructed test images, as well as between the original and defaced. The former metric represents the global correlation between two images, while the latter is aimed at predicting the perceived quality of a target image when compared to a reference image. We restrict our analyses to the front half of the images, since we are interested only in the face. It is important to highlight the difficulty in quantitatively evaluating the perceived realism of an image, since existing metrics may not closely track the nuances of the human visual and pattern recognition systems.

\begin{figure*}[htbp]
    \begin{minipage}[b]{0.45\linewidth}
        \centering
        \includegraphics[width=\linewidth]{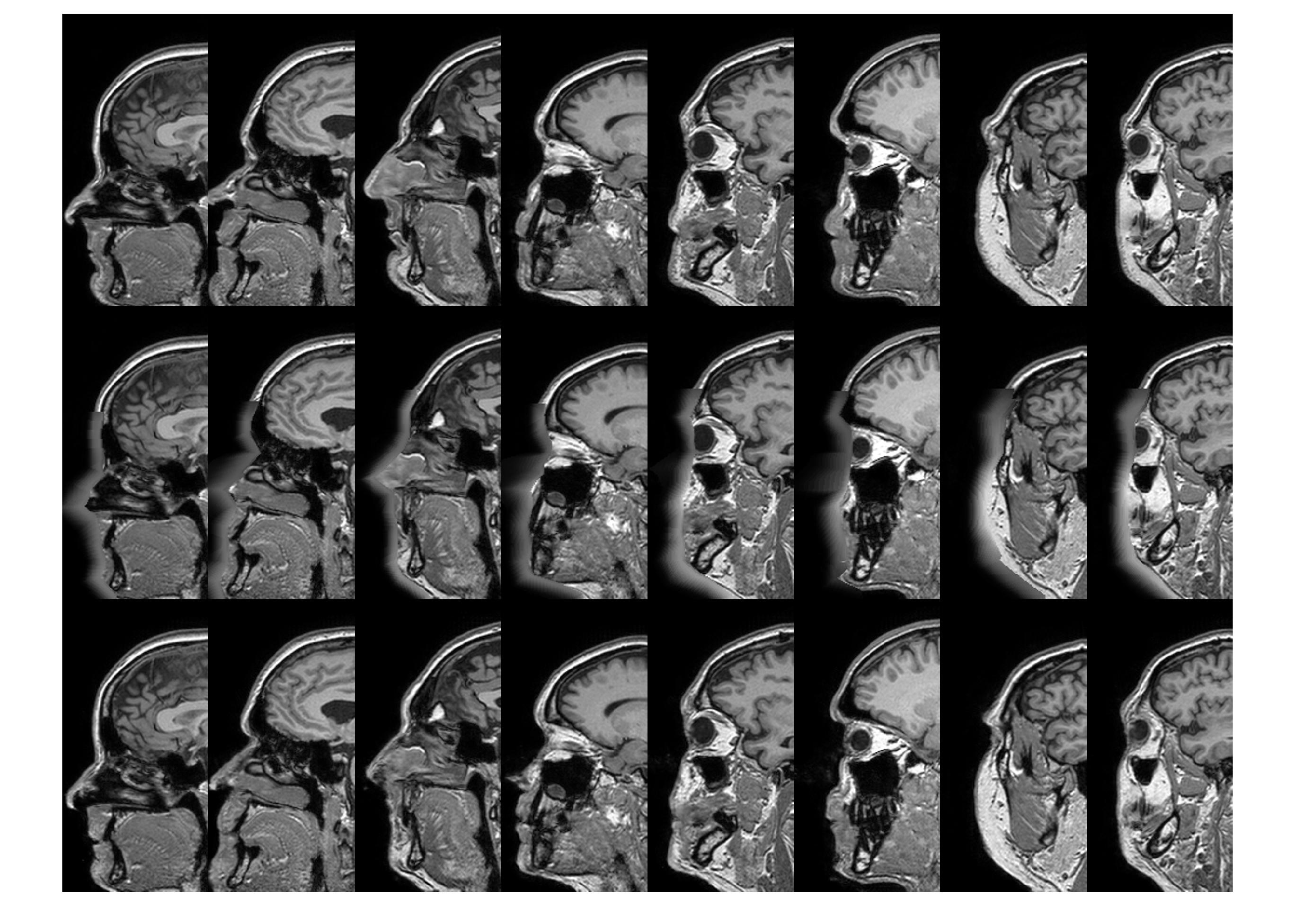}
    \end{minipage}
    \hfill
    \begin{minipage}[b]{0.45\linewidth}
        \centering
        \includegraphics[width=\linewidth]{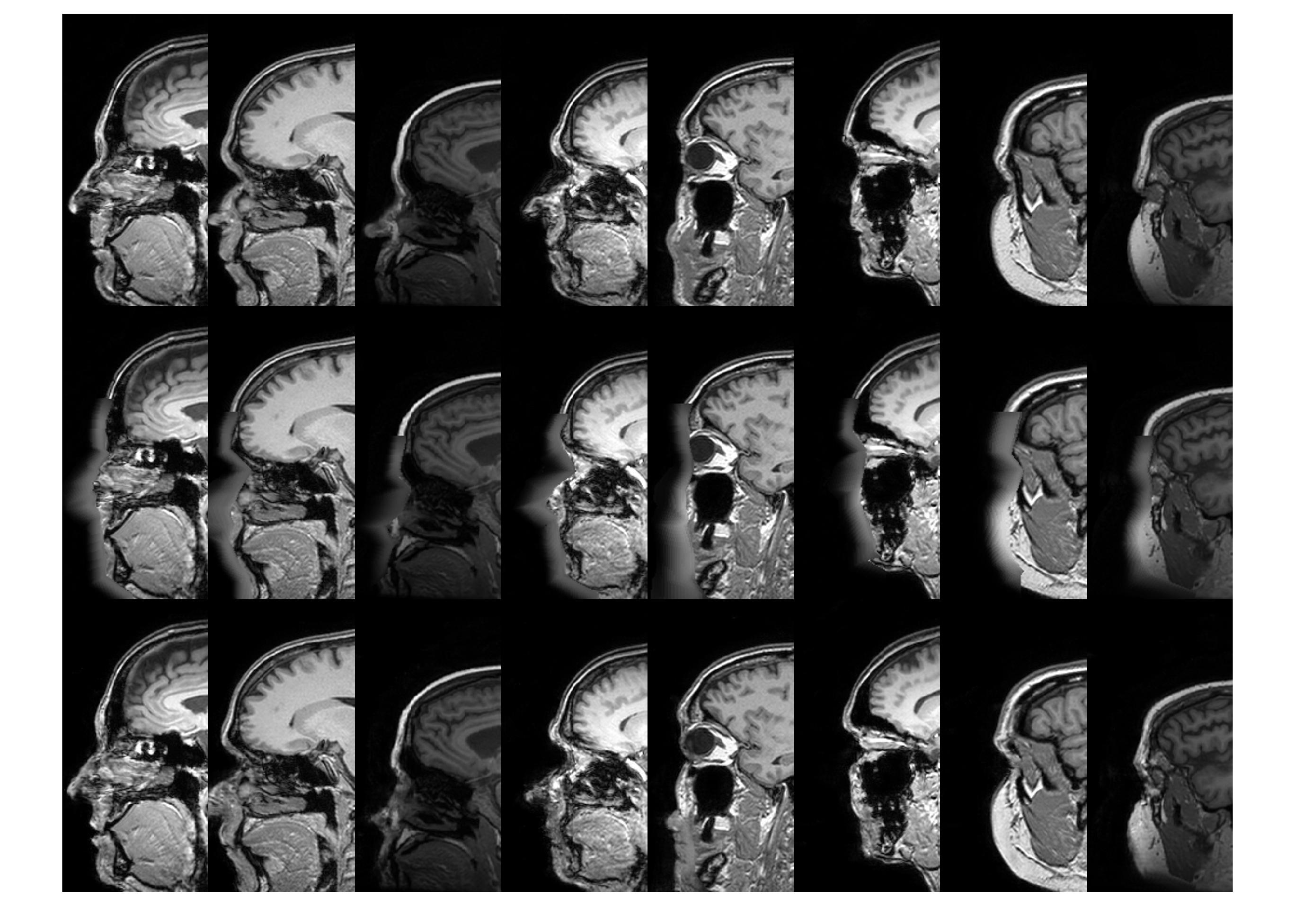}
    \end{minipage}
    \caption{Typical results of refacing face-blurred images. Left: results for training using only subjects from Guy's hospital, Right: results for training using data from all 3 sites. Top row: original image, middle row: face-blurred image, bottom row: reconstructed image. CycleGAN learns to perform a deconvolution, to reconstruct the anonymized face.}
    \label{fig:results_blurred}
\end{figure*}

\begin{figure*}[htbp]
    \begin{minipage}[b]{0.45\linewidth}
        \centering
        \includegraphics[width=\linewidth]{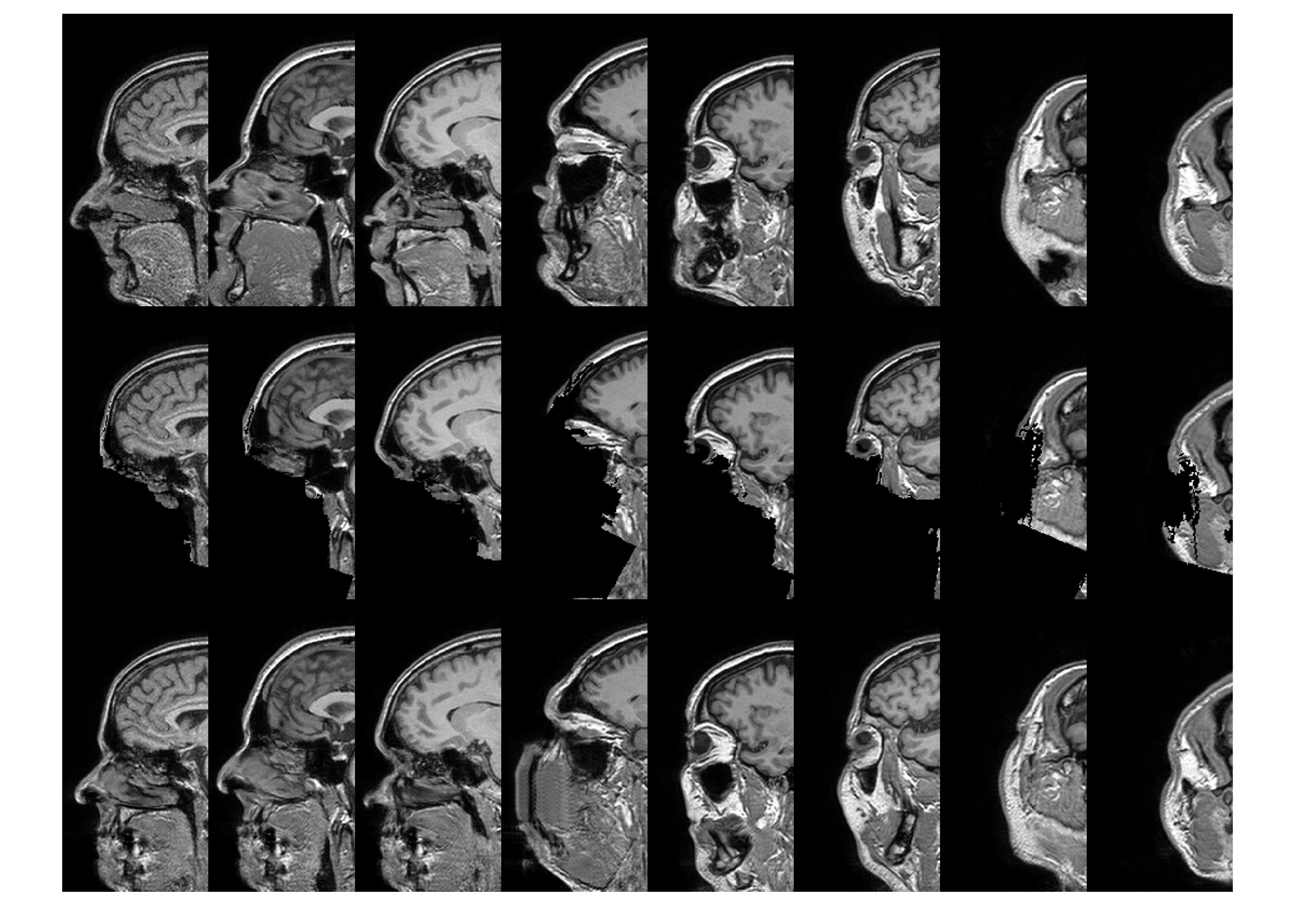}
    \end{minipage}
    \hfill
    \begin{minipage}[b]{0.45\linewidth}
        \centering
        \includegraphics[width=\linewidth]{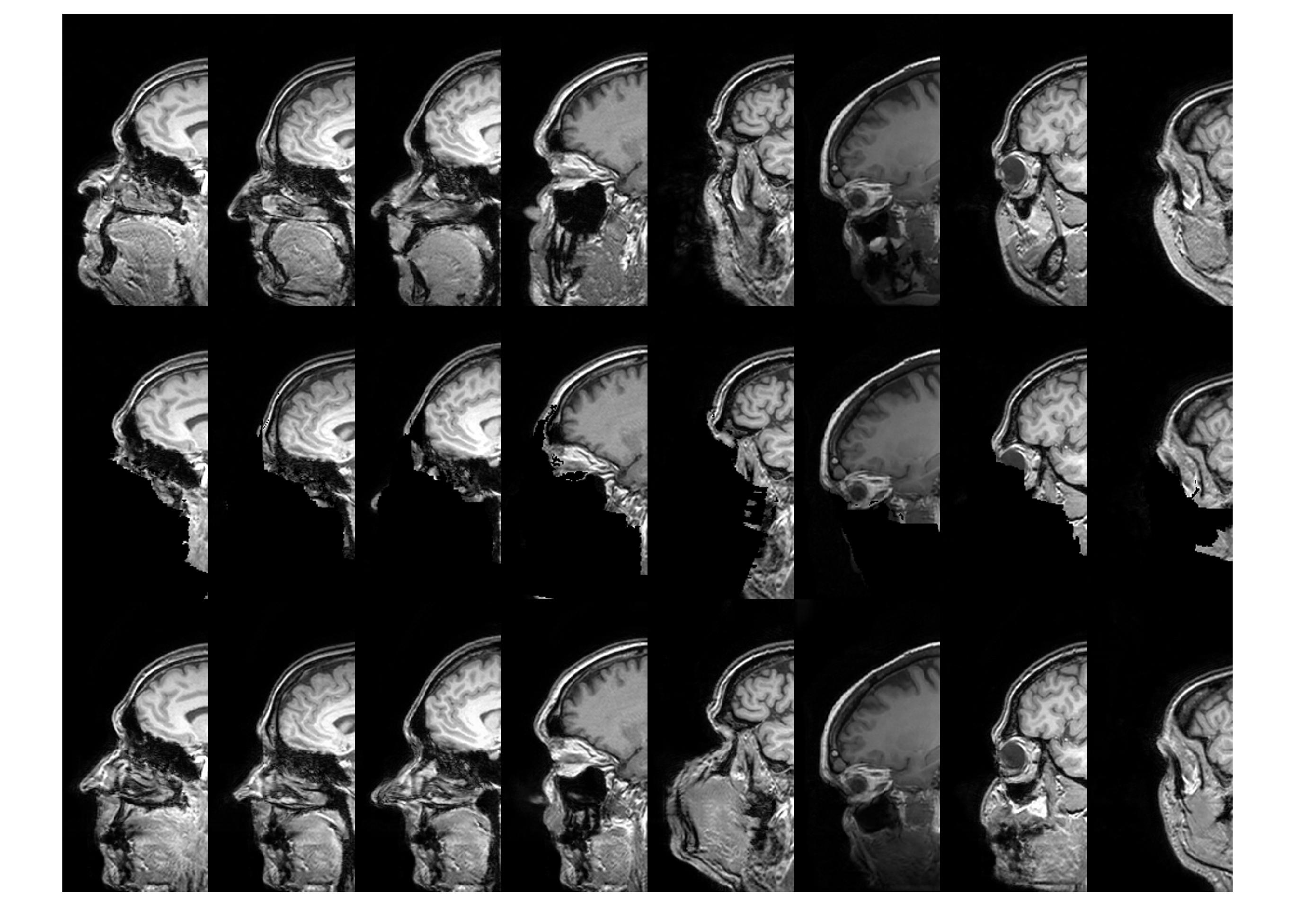}
    \end{minipage}
    \caption{Typical results of refacing face-removed images. Left: results for training using only subjects from Guy's hospital, Right: results for training using data from all 3 sites. Top row: original image, middle row: face-removed image, bottom row: reconstructed image. CycleGAN learns to add a face, but in many cases it is not the correct face.}
    \label{fig:results_removed}
\end{figure*}

\begin{figure*}[htbp]
    \begin{minipage}[b]{\linewidth}
        \centering
        \includegraphics[width=0.95\linewidth]{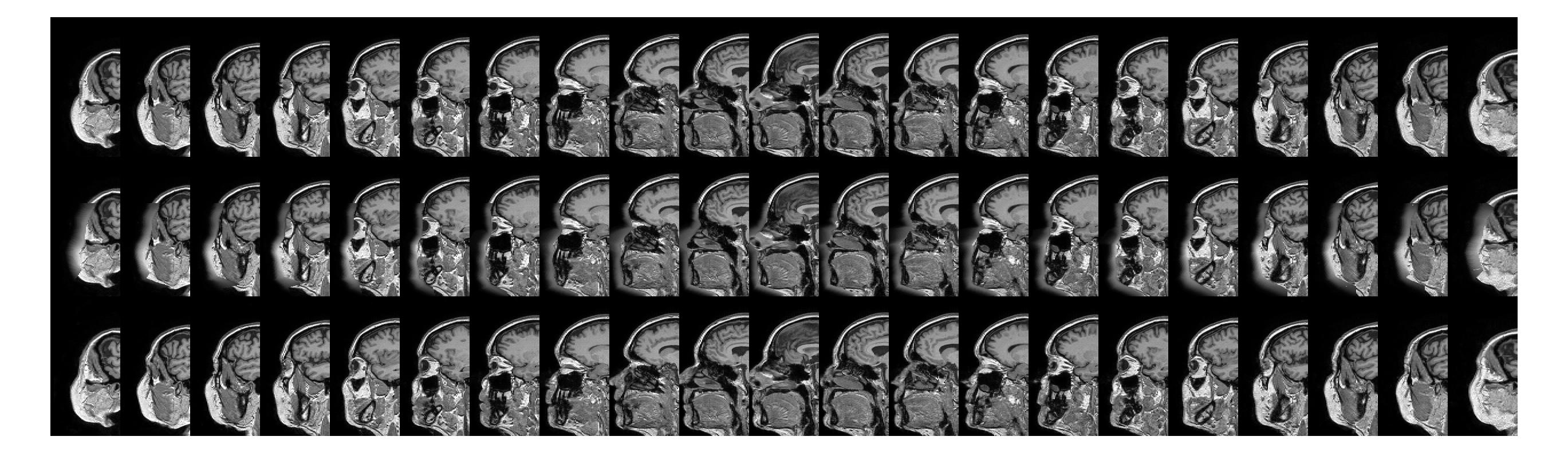}
    \end{minipage}
    \caption{Example results for all 21 slices of a test subject from the Guy's Hospital data. Good results are achieved for most slices. Top row: original image, middle row: face-blurred image, bottom row: reconstructed image.}
    \label{fig:slices}
\end{figure*}

\section{Results}
\subsection{Face blurring}
Our approach managed to convincingly reconstruct the facial features for the face-blurred images (see Figure~\ref{fig:results_blurred}). The results for the single dataset experiment were particularly consistent (see Figure~\ref{fig:slices}). Results were also positive for the full experiment, but showing a slight dependence of reconstruction quality on acquisition site. Qualitatively the images from Guy's Hospital attained the best results, followed by those of Hammersmith Hospital, and finally those of the Institute of Psychiatry. 

The quantitative results in Figures~\ref{fig:corr_ssim_guys}~and~\ref{fig:corr_ssim_full} show a high and approximately constant correlation and SSIM for both models. In all cases the differences between the mean metrics for the anonymized and the reconstructed images are very small. This, together with the fact that the metrics do not track the subjective variation in reconstruction quality for the three datasets, points to the difficulty in quantifying reconstruction quality.

\subsection{Face removal}
Only limited success was achieved for the face-removed images (see Figure~\ref{fig:results_removed}). While the GAN managed to restore a credible face in some cases, this rarely resembled the original face. The results also suffered from mode collapse, with particular image patches occurring in the output for many different input images. Another common problem is a sharp vertical cutoff on the front of the face, especially on the nose. This is due to the heads of many of the subjects used for training being cut against the boundaries of the volume

\begin{figure}[t!]
    \begin{minipage}[b]{0.88\linewidth}
        \centering
        \includegraphics[width=0.95\linewidth]{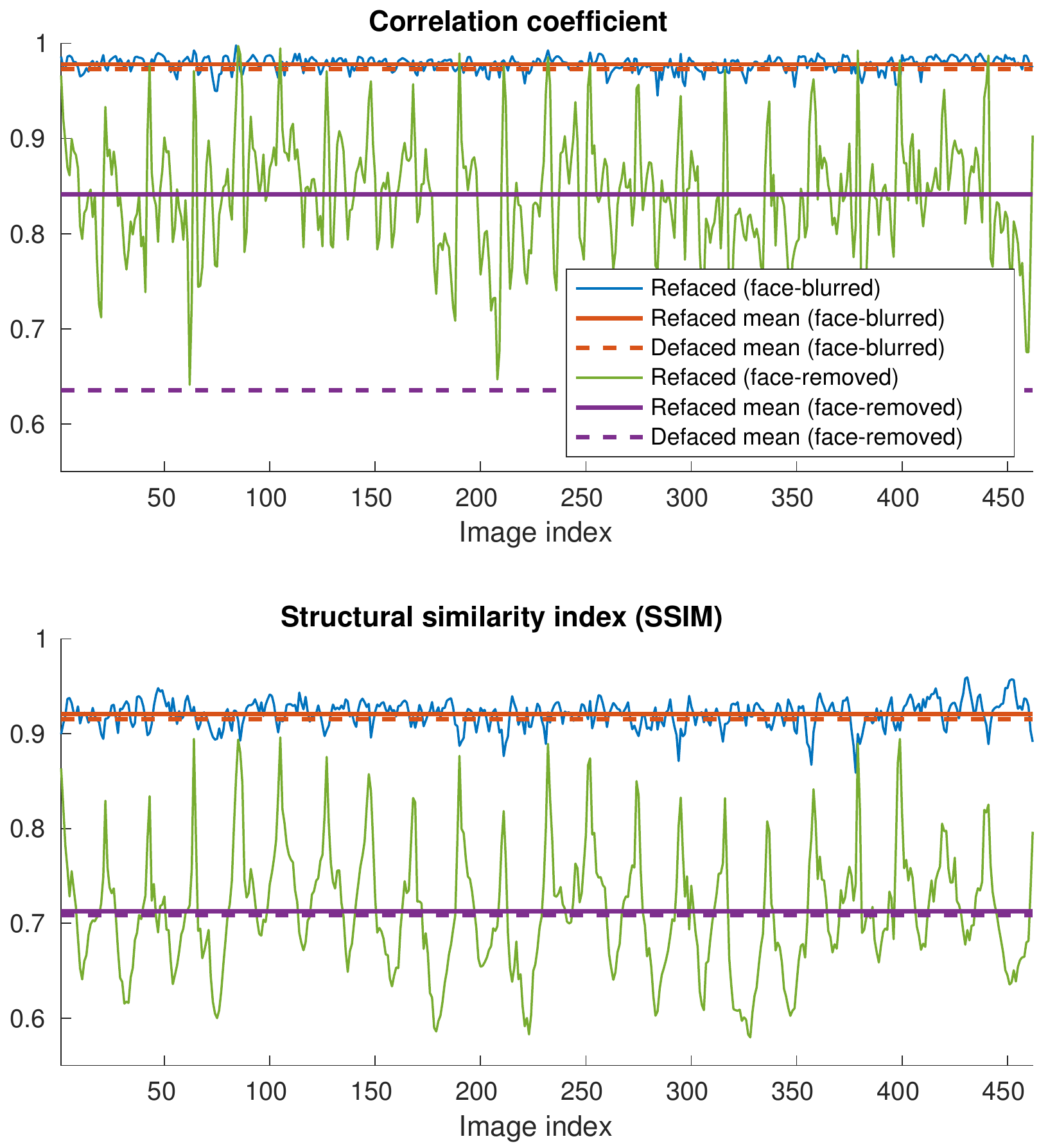}
    \end{minipage}
    \caption{Correlation and structural similarity between original and reconstructed test images after training using only subjects from Guy's hospital. As expected, it is easier to reconstruct face blurred images, compared to face removed images.}
    \label{fig:corr_ssim_guys}
\end{figure}

\begin{figure}[t!]
    \begin{minipage}[b]{0.88\linewidth}
        \centering
        \includegraphics[width=0.95\linewidth]{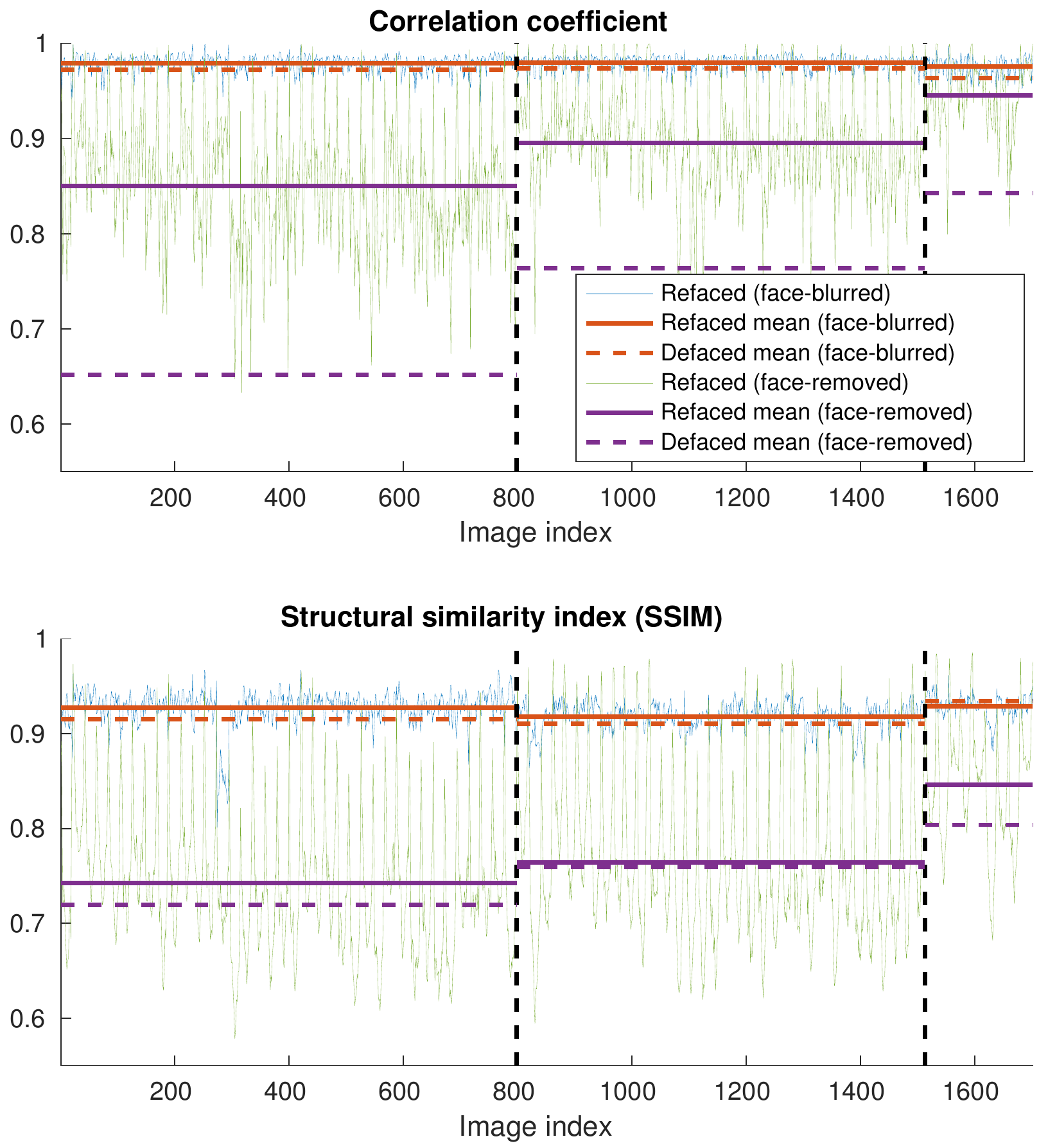}
    \end{minipage}
    \caption{Correlation and structural similarity between original and reconstructed test images after training using data from all 3 sites. Vertical lines separate images by acquisition site (left to right: Guy's Hospital, Hammersmith Hospital, Institute of Psychiatry). Means were calculated separately for images from each site.}
    \label{fig:corr_ssim_full}
\end{figure}

The reconstructed faces for the single dataset experiment are generally more shallow compared to the original data. Outcomes were slightly better for the full experiment, with a similar pattern of dependence on the acquisition site as seen in the face blurring case. Figures~\ref{fig:corr_ssim_guys}~and~\ref{fig:corr_ssim_full} show consistently worse quantitative results for the face-removed images compared to the face-blurred ones. Again, the differences in mean metrics across datasets do not match the qualitative evaluation. Mean correlation is the one metric that sees a significant improvement from refacing, compared to the defaced images.


\section{Discussion}
The successful reconstructions raise some concerns about the potential vulnerability of certain common anonymization methods used for MRI data. While the current implementation has only been shown to work on test data coming from the same dataset used to train the model, pre-trained networks are commonly used outside of their original domain. As an example, the Human Connectome Project \cite{VanEssen2013} provides face-blurred data of 1113 subjects, which could have been used to test the generalization properties of our procedure. However, to respect the privacy of these subjects, we have not attempted to apply our trained GANs on them. In addition, the generalization properties of the network could be improved by training it on datasets acquired from multiple sites and with different scanning parameters. 

Restoring facial features from blurred data requires that CycleGAN learns a deconvolution. Given the amount of information remaining in the image, such as some of the facial bones, and even the trajectory traced by the blurred face, this algorithm proved to be reversible to a significant extent. Restoring the complete face from zeroed out data poses a much more challenging inpainting problem. Even then, some success was achieved using an established GAN approach. In other domains, great success has been achieved by using dedicated inpainting architectures \cite{iizuka2017globally}, which can be another way to perform refacing.

In regards to the model employed, the choice to use a 2D GAN was made on the basis of constraints in available memory and processing time. When applied to every slice of a volume, such an approach would show discontinuities between contiguous slices. For this reason, no attempt was made to identify the subjects using volume renderings. Future work might examine the possibility of using a 3D GAN, which we expect would yield better results and solve the discontinuity problem. Another possible avenue for future investigation would be the use of a supervised learning model such as Pix2Pix \cite{Isola2017image}, since the available data is paired.

A potential legitimate application for refacing can be the improvement of morphometric estimates from anonymized data. It has been shown that even minimal anonymization procedures such as facial blurring can have an impact on morphometric estimates, such as subcortical volume and cortical thickness \cite{holmes2015brain}. Recovering the face of each subject could improve the correctness of these estimates.

\clearpage
\newpage
\bibliographystyle{IEEEbib}
\bibliography{refs}

\end{document}